\documentclass[letterpaper, 10 pt, conference]{sty/ieeeconf}

\IEEEoverridecommandlockouts
\usepackage{amsmath,amssymb,amsfonts,siunitx}
\usepackage{graphicx,epsfig,layouts}

\let\labelindent\relax
\usepackage{enumitem}
\usepackage{comment}
\usepackage{tabularx}

\usepackage[dvipsnames]{xcolor}
\usepackage[font=small,labelfont=bf]{caption}
\usepackage[position=b]{subcaption}
\usepackage[utf8]{inputenc}

\usepackage{flushend}   
\usepackage{bm}
\usepackage{tikz}
\usetikzlibrary{calc,math}
\usepackage{setspace}
\usepackage{algorithm}
\usepackage[noEnd=true,indLines=true,commentColor=black]{sty/algpseudocodex}
\usepackage{xurl}
\usepackage[hidelinks]{hyperref}
\usepackage{cleveref}
\usepackage{cite}
\usepackage{sty/custom}
\usepackage{eso-pic}

\title{\LARGE \bf D4orm: Multi-Robot Trajectories with\\ Dynamics-aware Diffusion Denoised Deformations}

\author{Yuhao Zhang, Keisuke Okumura, Heedo Woo, Ajay Shankar, Amanda Prorok%
\thanks{The authors are with the University of Cambridge, UK.
KO is also with National Institute of Advanced Industrial Science and Technology (AIST), Japan.
Emails: {\tt\scriptsize \{yz981,ko393,hw527,as3233,asp45\}@cst.cam.ac.uk}.}%
\thanks{This research was funded in part by the EPSRC funded INFORMED-AI project EP/Y028732/1 and in part by European Research Council (ERC) Project 949940 (gAIa). We gratefully acknowledge their support. KO was partially supported by JSPS Overseas Research Fellowship.}%
\thanks{Code and video: \textcolor{blue}{\texttt{https://github.com/proroklab/d4orm}}}}

\begin{document}

\newcommand{\boldslash}{%
  \leavevmode\rlap{/}\kern0.3pt/%
}
\AddToShipoutPictureFG*{%
  \AtPageUpperLeft{%
    \hspace*{\dimexpr 1in+\oddsidemargin\relax}%
    \raisebox{-0.5in}{%
      \begin{minipage}{\textwidth}
        \footnotesize
        \fontfamily{phv}\selectfont
        \textbf{\copyright~2025 IEEE -- Accepted to IEEE{\boldslash}RSJ International Conference on Intelligent Robots and Systems (IROS) 2025}
      \end{minipage}%
    }%
  }%
}

\maketitle
\begin{abstract}
This work presents an optimization method for generating kinodynamically feasible and collision-free multi-robot trajectories that exploits an incremental denoising scheme in diffusion models. Our key insight is that high-quality trajectories can be discovered merely by denoising noisy trajectories sampled from a distribution. This approach has no learning component, relying instead on only two ingredients: a dynamical model of the robots to obtain feasible trajectories via rollout, and a fitness function to guide denoising with Monte Carlo gradient approximation. The proposed framework iteratively optimizes a deformation for the previous trajectory with the current denoising process, allows \textit{anytime} refinement as time permits, supports different dynamics, and benefits from GPU acceleration. Our evaluations for differential-drive and holonomic teams with up to 16 robots in 2D and 3D worlds show its ability to discover high-quality solutions faster than other black-box optimization methods such as MPPI. In a 2D holonomic case with 16 robots, it is almost twice as fast. As evidence for feasibility, we demonstrate zero-shot deployment of the planned trajectories on eight multirotors.
\end{abstract}

{
\newcommand{\entry}[5]{
  \tikzmath{
    \x = #1 * \xsize;
    \y = - #2 * \ysize;
  }
  \node[] at (\x, \y) {\includegraphics[width=0.18\linewidth,trim={2 2 2 2},clip]{images/denoising/#3/diffusion_step_#4_#5}};
}
\begin{figure}[!hp]
\centering
\begin{tikzpicture}
\node[anchor=north east] at (0, -0.01) {\includegraphics[width=1\linewidth,trim={0 2.5cm 0 0},clip]{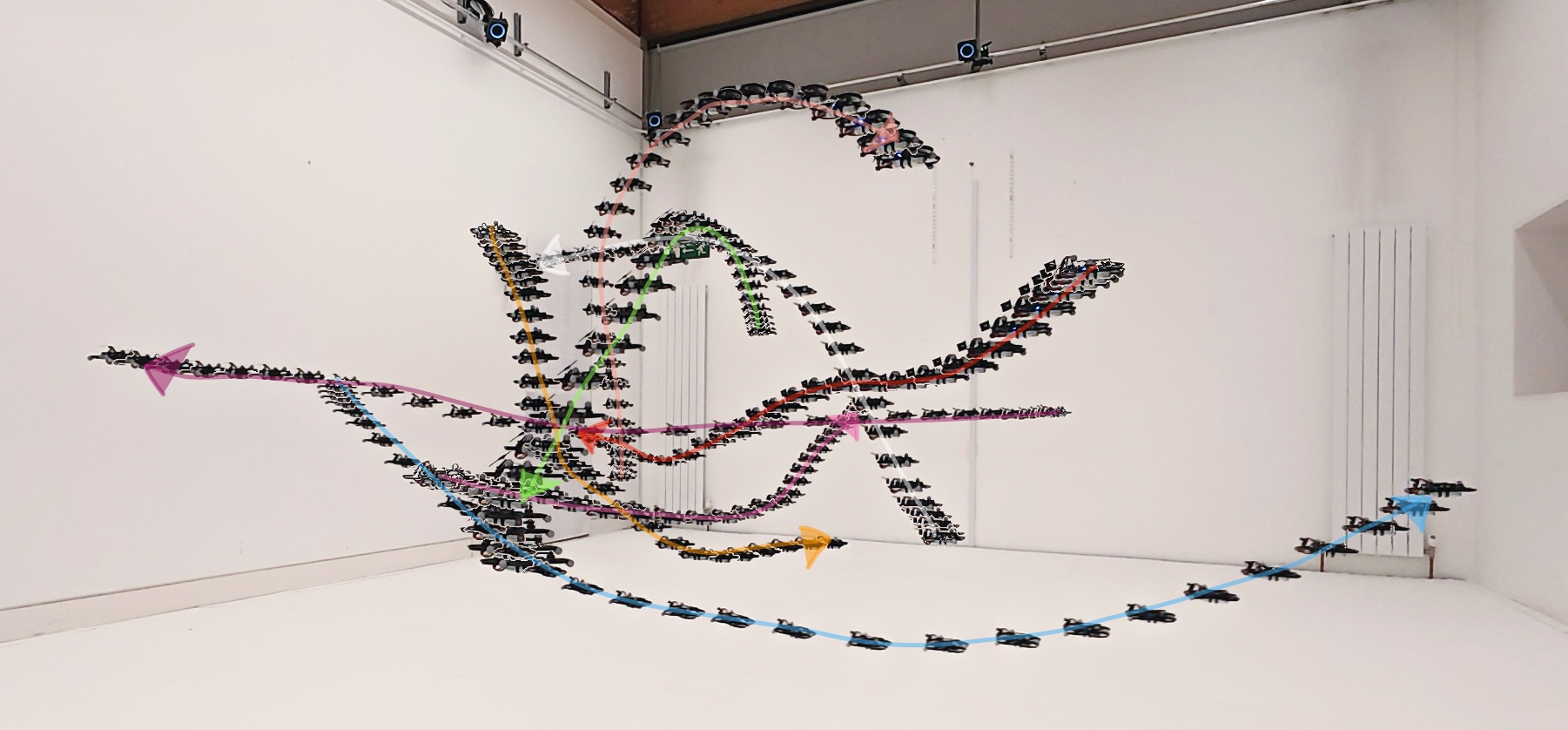}};
\node[anchor=north east] at (0, 0)  {\includegraphics[width=0.25\linewidth]{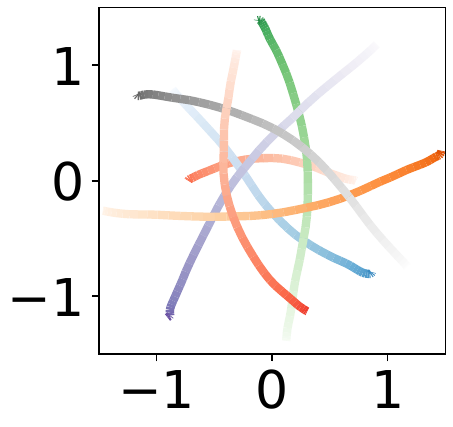}};
\tikzmath{
  \y = -3.8;
}
\node[anchor=north east] at (0, \y)  {\includegraphics[width=0.95\linewidth]{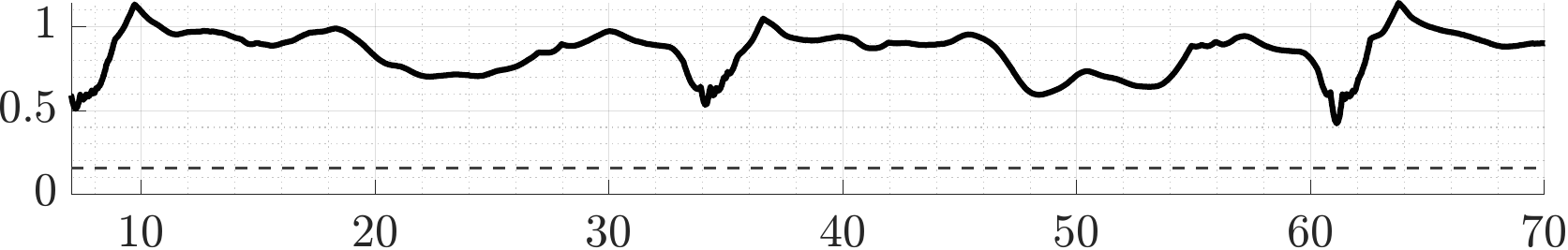}};
\node[fill=white,inner sep=1pt] at (-0.7, \y-0.97) {\scriptsize Time [s]};
\node[rotate=90] at (-8.5, \y-0.7) {\scriptsize Dist [m]};
\end{tikzpicture}\medskip\\
\begin{tikzpicture}
  \tikzmath{
    \xsize = 1.6;
    \ysize = 1.6;
  }
  \entry{0}{0}{2dholo}{1}{0}
  \entry{1}{0}{2dholo}{1}{24}
  \entry{2}{0}{2dholo}{1}{49}
  \entry{3}{0}{2dholo}{1}{74}
  \entry{4}{0}{2dholo}{1}{99}
  \entry{0}{1}{2dholo}{2}{0}
  \entry{1}{1}{2dholo}{2}{24}
  \entry{2}{1}{2dholo}{2}{49}
  \entry{3}{1}{2dholo}{2}{74}
  \entry{4}{1}{2dholo}{2}{99}
  \entry{0}{2}{2dholo}{3}{0}
  \entry{1}{2}{2dholo}{3}{24}
  \entry{2}{2}{2dholo}{3}{49}
  \entry{3}{2}{2dholo}{3}{74}
  \entry{4}{2}{2dholo}{3}{99}
  \entry{0}{3}{2dholo}{4}{0}
  \entry{1}{3}{2dholo}{4}{24}
  \entry{2}{3}{2dholo}{4}{49}
  \entry{3}{3}{2dholo}{4}{74}
  \entry{4}{3}{2dholo}{4}{99}
  \entry{0}{4}{2dholo}{5}{0}
  \entry{1}{4}{2dholo}{5}{24}
  \entry{2}{4}{2dholo}{5}{49}
  \entry{3}{4}{2dholo}{5}{74}
  \entry{4}{4}{2dholo}{5}{99}
  %
  \entry{0}{5.0}{3dholo}{1}{0}
  \entry{1}{5.0}{3dholo}{1}{24}
  \entry{2}{5.0}{3dholo}{1}{49}
  \entry{3}{5.0}{3dholo}{1}{74}
  \entry{4}{5.0}{3dholo}{1}{99}
  %
  \entry{0}{6.0}{2dcar}{1}{0}
  \entry{1}{6.0}{2dcar}{1}{24}
  \entry{2}{6.0}{2dcar}{1}{49}
  \entry{3}{6.0}{2dcar}{1}{74}
  \entry{4}{6.0}{2dcar}{1}{99}
  %
  \node[] at (\xsize * 2, 1.2) {denoising step $\rightarrow$};
  \node[rotate=90] at (-1.2, -\ysize * 2) {$\leftarrow$ iteration};
  {
    \scriptsize
    \node[] at (\xsize * 0, 0.9) {99};
    \node[] at (\xsize * 1, 0.9) {75};
    \node[] at (\xsize * 2, 0.9) {50};
    \node[] at (\xsize * 3, 0.9) {25};
    \node[] at (\xsize * 4, 0.9) {0};
    \node[] at (-0.9, -\ysize * 0) {1};
    \node[] at (-0.9, -\ysize * 1) {2};
    \node[] at (-0.9, -\ysize * 2) {3};
    \node[] at (-0.9, -\ysize * 3) {4};
    \node[] at (-0.9, -\ysize * 4) {5};
    \node[] at (-0.9, -\ysize * 5.0) {1};
    \node[] at (-0.9, -\ysize * 6.0) {1};
  }
\end{tikzpicture}
\caption{
  Zero-shot deployment of deconflicted trajectories with eight multirotors (top) with denoised trajectories (top-right inset); the minimum pairwise inter-robot distance shows safety margin w.r.t robot radius (solid and dotted black lines).
  The lower plots visualize denoising of 16 2D-holonomic, 8 3D-holonomic, and 8 differential-drive robots. Each plot shows the result of adding deformation denoised at current step to trajectory given by previous iteration.
}
\label{fig:denoising}
\end{figure}
}

\section{Introduction}
Generating conflict-free state-to-state trajectories for robot teams is a critical task when operating multiple robots in a shared workspace, and is frequently required in areas such as warehouse automation~\cite{wurman2008coordinating} and transportation systems~\cite{dresner2008multiagent}.
In such scenarios, robots need to operate with tight coordination while minimizing redundant movements to optimize various metrics of system performance (such as flowtime).
This is achieved by optimizing trajectory plans for the entire team.

While the quality of collision-free trajectories is easily specified, the corresponding ``joint’’ optimization problem, where all robots' states are considered jointly, is often unsuitable for classical numerical methods.
This is primarily due to the size of the joint configuration space, which grows exponentially with the number of robots~\cite{hopcroft1984complexity}.
Furthermore, the optimization landscape is generally non-convex, can contain multiple equivalent solutions, and has constraints that make it difficult to compute exact analytical gradients.
Recent trends therefore relax the objective of synthesizing globally optimal trajectories by decoupling robot-wise states from the joint representation, while often choosing suboptimal and conservative actions due to an incomplete state representation~\cite{chen2015decoupled,luis2019trajectory,tordesillas2021mader,zhou2021ego,csenbacslar2023rlss}.
Although their advances are remarkable and their decentralization possibilities appealing, the lack of coordination guarantees makes them unsuitable for safety-critical multi-robot platforms, which could be the future infrastructure that underpins our daily lives.
With these in mind, reliable and scalable multi-robot trajectory optimization methods in the joint representation remain a pivotal technology.

In this work, we investigate a novel optimization mechanism that generates collision-free and kinodynamically feasible trajectories in the joint representation.
We sidestep each of the aforementioned limitations of classical numerical methods by employing strategies from diffusion models, which have successfully captured high-dimensional features in various domains, originally in image generation~\cite{ho2020denoising,rombach2022high}, but also in robot motion generation~\cite{chi2023diffusion}.
Specifically, building on \emph{model-based diffusion}~\cite{pan2024model}, we propose \textsc{D4orm} (``deform"), that iteratively denoises joint control-space trajectories for the entire team by using diffusion denoising as a black-box (gradient-free) optimization tool.
Each denoising step refines a \textit{deformation} from the previous step's solution, and is guided by an on-the-fly Monte Carlo score approximation combined with an interpretable fitness function.
Unlike the traditional use of diffusion models, this dynamics-driven approach does not require supervised data and therefore support different robot dynamics.
The proposed framework is an \textit{anytime} algorithm, and benefits from GPU parallelization of Monte Carlo rollouts.

Illustrated as snapshots in \cref{fig:denoising}, our key insight is that, starting from noisy/infeasible trajectories, high-quality trajectories are discovered entirely through denoising, a process that only needs a model of the robot dynamics and a fitness (reward) function.
We present a range of studies in terms of computation time, success rate, and solution quality, with teams of 8-16 robots navigating in a dense obstacle-free workspace towards antipodal points on a circle, and with three distinct dynamics.
Our method retrieves deconflicted trajectories in these high-dimensional, potentially non-convex solution spaces within reasonable deadlines while outperforming baselines.
For instance, we show planning for 16 2D-holonomic robots in approx. \SI{2}{\second} on average over a planning horizon of 100 steps, which is near $2\times$ faster than other sampling-based optimization methods.
Finally, we deploy a team of eight multirotors in a zero-shot manner, proving the kinodynamic feasibility of generated trajectories.

\section{Related Work}

There exists a rich literature on multi-agent and multi-robot trajectory optimization.
Instead of decoupled approaches that perform an optimization process locally per-robot, e.g.,~\cite{chen2015decoupled,luis2019trajectory,tordesillas2021mader,zhou2021ego,csenbacslar2023rlss}, we are interested in improving the performance of coupled approaches that obtain coordinated trajectories.
These methods are often categorized as either \emph{sampling-based motion planning (SBMP)} or \emph{numerical optimization}.
The former, SBMP, uses discrete combinatorial search over a roadmap approximation of the joint configuration space, constructed by random sampling~\cite{vsvestka1998coordinated,solovey2016finding,okumura2023quick}.
The latter reduces deconfliction to a numerical optimization formulation and then typically applies well-established solvers to obtain a solution~\cite{augugliaro2012generation,kushleyev2013towards,adajania2023amswarm}.
Both categories suffer from the curse of dimensionality as the number of agents increases.

Our proposed denoising method belongs to the orthogonal category of what we call \emph{sampling-based optimization} methods, which, unlike SBMP, aim to sample a solution trajectory from a high-dimensional space, by massive sampling attempts.
Unlike numerical optimization and SBMP, sampling-based optimization methods have an easily accelerated structure by parallelizing this trajectory collection with GPUs.
While such methods have been successfully applied to single-robot control in challenging environments~\cite{williams2018information}, applying it to multi-robot systems also faces the same dimensionality issue.
In fact, many existing sampling-based optimization methods for multi-robot control use decoupled representations~\cite{streichenberg2023multi,trevisan2024biased,jiang2024distributed}.
In contrast, we aim to sample the joint trajectories directly through diffusion denoising, a process more equipped to handle high dimensionality.

Diffusion models, which originally received considerable attention in image generation~\cite{ho2020denoising,rombach2022high}, are a popular choice for learning-based robot motion generation~\cite{janner2022planning,chi2023diffusion}.
It is chiefly characterized by its ability to extract features in very high-dimensional spaces, such as those that represent an image, or multi-robot motion.
Indeed, diffusion has been applied in multi-agent use-cases such as in trajectory prediction~\cite{jiang2023motiondiffuser}, or motion planning combined with constrained optimization~\cite{shaoul2025multi,liang2024multi}.
While such diffusion-based generative methods aim to imitate demonstration trajectories, recent work has also proposed learning-free trajectory generation via Monte Carlo gradient approximation, called model-based diffusion (MBD)~\cite{pan2024model}.
Our work builds on MBD, explained in detail in \cref{sec:method}, which was originally designed for single-robot planning.
We expand it to multi-robot settings with the introduction of team-level fitness (reward) functions plus a key methodological innovation of iteratively optimizing \textit{deformation} vectors to solve more challenging problems.

\section{Problem Definition}
\label{sec:problem-def}
In the following, we use a braced superscript $\{k\}$ to denote robot indices, and a plain subscript $i$ for a denoising step.

A target system consists of a team of $n$ homogeneous spherical robots $R = \{1, 2, \ldots n \}$ each with a radius $\mathrm{R}_a$.
Their dynamics are each governed by $\dot{x} = f_{\mathrm{dyn}}(x, u)$, where
$x \in \mathcal{X} \subset \mathbb{R}^{d_x}$ and
$u \in \mathcal{U} \subset \mathbb{R}^{d_u}$
denote state and control vectors, respectively.
Given their joint initial and terminal states, $\mathcal{S}, \mathcal{T} \in \mathcal{X}^n$,
the \textbf{objective} is to generate a list, $\tau$, representing a set of $n$ collision-free and kinodynamically feasible trajectories.
Specifically, for robots $k, l \in R$, $\tau^{\{k\}} \in (\mathcal{X} \times \mathcal{U})^{H}$ represents the trajectory for the $k$-th robot over some finite horizon $H$ as a sequence of state-control pairs, sampled at time intervals $\triangle t \in \mathbb{R}_{>0}$, which satisfies:
\begin{alignat}{2}
    \tau^{\{k\}}[t+1] &= \texttt{RK4}(\tau^{\{k\}}[t], f_{\mathrm{dyn}}, \triangle t)
      &&\quad \text{(feasibility)} \nonumber \\
    \tau^{\{k\}}[1].x &= \mathcal{S}^{\{k\}}
      &&\quad \text{(init. cond.)} \nonumber \\
    \tau^{\{k\}}[H].x &= \mathcal{T}^{\{k\}}
      &&\quad \text{(term. cond.)} \nonumber \\
    \text{Dist}(\tau^{\{k\}}[t] &, \tau^{\{l\}}[t] ) > 2\cdot\mathrm{R}_a
      &&\quad \text{(safety)}
    \label{eqn:problem-conditions}
\end{alignat}
where $\texttt{RK4}$ denotes the fourth-order Runge-Kutta integration of system dynamics, and $\text{Dist}(\cdot)$ denotes the Euclidean distance between the position components of two trajectory states.
The quality, i.e., the reward to be maximized, of a feasible solution is evaluated in relation to the total travel time, represented in the following form:
\begin{equation}
\frac{1}{nH}
  \sum_{k \in R}
  \sum_{t=0}^{H}
   \mathrm{Ind}\left[ \tau^{\{k\}}[t].x = \mathcal{T}^{\{k\}} \right]
\label{eq:objective}
\end{equation}
where $\mathrm{Ind}[\cdot] = 1$ if the condition is true; zero otherwise.

\section{Deconflicting with Diffusion Denoising}
\label{sec:method}
We now describe the denoising optimization process that obtains a solution to the problem defined in \cref{eqn:problem-conditions}.
While finding an optimal solution is non-trivial due to the high dimensionality of the space, it is relatively straightforward to evaluate candidate solutions based on a fitness (reward) function.
Our approach, therefore, is to use rewards observed in a batch of candidate rollouts generated using a given $f_{\mathrm{dyn}}$ to iteratively refine new candidates.
We begin by explicitly defining such a reward function that represents \cref{eq:objective} for multi-robot navigation scenarios. We use the term `reward' to be consistent with \textit{model-based diffusion}\cite{pan2024model}.

\subsection{Reward Structure For Multi-Robot Trajectories}
We define the general multi-robot reward as comprised of two parts: a quality of goal navigation reward, $r\sub{goal}$, and a reward for inter-robot safety, $r\sub{safe}$.
Using a weighting parameter $w_t \in \mathbb{R}$, our reward for the joint trajectory $\tau$ over the horizon $H$ is expressed as
\begin{align}
  r(\tau) = \frac{1}{nH} \sum_{t=1}^{H} \sum_{k \in R} \Big( r_\text{goal}(\tau^{\{k\}}, t) + w_t \cdot r_\text{safe}(\tau^{\{k\}}, t) \Big).
  \label{eqn:generic-reward}
\end{align}

The first term optimizes the $k$-th trajectory \(\tau^{\{k\}}\) based on the objective function in \cref{eq:objective}, but with a dense reward structure, while the second term explicitly penalizes collisions.
Specifically, in our setting, given a target position \( p^{\{k\}}_\mathcal{T} \in \mathcal{T}^{\{k\}} \),
\begin{align}
r_{\text{goal}}(\tau^{\{k\}}, t) &= 1 - \frac{\| p^{\{k\}}[t] - p_\mathcal{T}^{\{k\}} \|}{\| p^{\{k\}}[1] - p_\mathcal{T}^{\{k\}} \|}\\
r_{\text{safe}}(\tau^{\{k\}}, t) &=
    \begin{cases}
        -1,  & \text{if } \| p^{\{k\}}[t] - p^{\{l\}}[t] \| \leq 2\mathrm{R}_a + \epsilon \\
        0,   & \text{otherwise}
    \end{cases}
\end{align}
where $\epsilon \in \mathbb{R}_{+}$ defines a safety margin, and $l \in R\setminus k$.

The simple reward design from \cref{eqn:generic-reward} guides the denoising process which we describe in the following subsections.

\subsection{Diffusion Denoising without Data}
We begin by providing a general overview of diffusion and denoising, adapted here for completeness from prior work on model-based diffusion~\cite{pan2024model}.
For now, consider a single-robot scenario, i.e., $\tau$ as a \textit{single robot} trajectory $\tau \in (\mathcal{X} \times \mathcal{U})^H$.
In the scheme of diffusion models, we are interested in sampling a solution from a target distribution $p_0$ that assigns a high density to a solution with higher rewards.
Using a temparture parameter $\lambda \in \mathbb{R}_{>0}$, we represent $p_0$ as:
\begin{align}
  p_0(\tau) \propto \exp\left(\frac{r(\tau)}{\lambda}\right)
\end{align}

Sampling a solution directly from $p_0$ is, however, significantly challenging because $\tau$ lies in a high-dimensional space, e.g., $H(d_x + d_u)$ for the single-robot case.
Diffusion models are a successful framework for dealing with this problem through an iterative denoising process.
The forward process slowly corrupts the structure of the data by adding noise.
Mathematically, each sample $\tau_i$ is corrupted sequentially from $p_0$ to an isotropic Gaussian $p_N$ with schedule parameters $\alpha_t$, using Gaussian noise of
\begin{align}
  p_{i|i-1}(\tau_t | \tau_{i-1}) = \mathcal{N}\left(\sqrt{\alpha_i}\tau_{i-1}, (1 - \alpha_i) I\right)
\end{align}
The backward process $p_{i-1|i}(\cdot)$ is the reverse of the forward process $p_{i|i-1}(\cdot)$.
When this backward distribution is available, we could reconstruct $p_0$ through:
\begin{align}
  p_{i-1}(\tau_{i-1}) &= \int p_{i-1|i} (\tau_{i-1} | \tau_{i}) p_i(\tau_i) d\tau_i
  \\
  p_0(\tau_0) &= \int p_N(\tau_N) \prod_{i=N}^1 p_{i-1|i}(\tau_{i-1} | \tau_i) d\tau_{1:N}
\end{align}

Traditional diffusion models solve the backward process by learning a score function from the data.
Meanwhile, the model-based diffusion (MBD)~\cite{pan2024model} employs Monte Carlo score estimation.
Using the inverse scale of the forward process, MBD employs the denoising process with
\begin{align}
\tau_{i-1} = \frac{1}{\sqrt{\alpha_i}} \left( \tau_i + (1 - \bar{\alpha}_i) \nabla_{\tau_i} \log p_i(\tau_i) \right)
\label{eq:denoising}
\end{align}
where $\bar{\alpha}_i = \prod_{j = 1}^i \alpha_j$.
The score term is approximated by
\begin{align}
  \nabla_{\tau_i} \log p_i(\tau_i) \approx
  -\frac{\tau_i}{1 - \bar{\alpha}_i}
  + \frac{\sqrt{\bar{\alpha}_i}}{1 - \bar{\alpha}_i}\bar{\tau}
\label{eq:guidance}
\end{align}
Here $\bar{\tau}$ is a weighted average of samples around $\tau_i$, based on the target distribution $p_0$:
\begin{align}
\Gamma \sim \mathcal{N}\left( \frac{\tau_i}{\sqrt{\bar{\alpha}_i}}, \left(\frac{1}{\bar{\alpha}_i} - 1\right) I \right),\;
\bar{\tau} = \frac{\sum_{\tau \in \Gamma} p_0(\tau) \tau}{\sum_{\tau \in \Gamma} p_0(\tau) }
\label{eq:mc}
\end{align}
Taking \cref{eq:denoising,eq:guidance,eq:mc} together, the one-step denoising in MBD is simply summarised as:
\begin{align}
\tau_{i-1} = \sqrt{\bar{\alpha}_{i-1}}\cdot\bar{\tau}
\label{eq:mbd-denoising}
\end{align}
In contrast to other sampling-based methods like MPPI~\cite{williams2018information}, which performs a softmax update, MBD introduces an extra noise schedule to both the sampling and update phases.

{
\begin{algorithm}[t!]
\caption{Denoising Optimization Process}
\label{algo:mbd}
\begin{algorithmic}[1]
\small
\State $U_N \sim \mathcal{N}(\bm{0}, I)$
\label{algo:mbd:init}
\Comment{initialize control trajectory}
\For{$i \gets N~\text{to}~1$}
\State Sample $M$ control trajectories:
\begin{align*}
\Gamma_u \sim
  \mathcal{N} \left( \frac{U_i}{\sqrt{\bar{\alpha}_i}}, \left( \frac{1}{\bar{\alpha}_i} - 1 \right) I \right)
\end{align*}\;
\label{algo:mbd:sample}
\State Sample $M$ trajectories:
$\Gamma \leftarrow \texttt{rollout}\left(\mathcal{S}, \Gamma_u \right)$
\label{algo:mbd:rollout}
\State Compute Monte Carlo estimation:
\begin{align*}
\bar{U} \leftarrow \frac
{\sum_{\tau \in \Gamma} p_0(\tau) (\tau.u)}
{\sum_{\tau \in \Gamma} p_0(\tau)}, \:
p_0(\tau) \approx \exp\left(\frac{r(\tau)}{\lambda}\right)
\end{align*}
\label{algo:mbd:mc}
\State Perform one-step denoising:
$U_{i-1} \leftarrow \sqrt{\bar{\alpha}_{i-1}} \: \bar{U}$
\label{algo:mbd:denoising}
\EndFor
\State \Return $\texttt{rollout}\left( \mathcal{S}, U_0 \right)$
\label{algo:mbd:final}
\end{algorithmic}
\end{algorithm}
}

\subsection{Denoising for Trajectory Optimization: \textsc{D4orm} Basics}
Now that we have established the process for denoising, the remainder of this section describes how we apply it for optimization.
Keeping in line with the diffusion terminology, we will slightly abuse the term `noisy trajectory' to mean a trajectory of states/controls sampled from a distribution other than the target distribution.
We note that the kinodynamic feasibility condition forces states and controls to be consistent using a given $f_{\mathrm{dyn}}$.
Thus, sampling them independently is impractical since the distribution approaches a Dirac delta function.
Therefore, similar to MBD, we sample control trajectories instead, and then retrieve state-control trajectories with rollout from the initial state, i.e, iteratively generate a state sequence with $x[t+1] = \texttt{RK4}(x[t], u[t], f_{\mathrm{dyn}}, \triangle t)$ given a control sequence.
This process on the trajectory batch $\Gamma$ can be effectively parallelized using GPUs.

\Cref{algo:mbd} describes how MBD actually solves the trajectory optimization.
Starting from a noisy control trajectory (Line~\ref{algo:mbd:init}), $U_N \in \mathcal{U}^{H}$ for single-robot, MBD first samples $M$ control trajectories surrounding the previous iteration (Line~\ref{algo:mbd:sample}).
These controls are then converted into feasible trajectories (Line~\ref{algo:mbd:rollout}) and used to approximate the score term (Line~\ref{algo:mbd:mc}), followed by an update from \cref{eq:mbd-denoising} (Line~\ref{algo:mbd:denoising}).
The denoised variable constitutes a solution (Line~\ref{algo:mbd:final}).
In practice, rewards are normalized within a batch to stabilize the denoising process.

MBD was originally developed for single-robot trajectory optimization.
Inspired by its ability to synthesize trajectories in high-dimensional space, the remaining part leverages it for multi-robot trajectory deconfliction.
We first extend the control trajectory, assumed in the previous description to be $U_i \in \mathcal{U}^{H}$, to the joint control trajectory for all robots, i.e. $U_i \in \mathcal{U}^{nH}$.
The rollout at Line~\ref{algo:mbd:rollout} is then performed for $n$ robots, and produces a batch of joint trajectories, each constituting $n$ robot trajectories $\tau \in (\mathcal{X} \times \mathcal{U})^{nH}$.
The rest of the denoising process operates in this joint trajectory representation to derive a solution $\tau$ to the multi-robot trajectory formulation in \cref{sec:problem-def}.

\subsection{Iterative Optimization of Deformations: \textsc{D4orm} Core}
Since finding solutions to multi-robot deconfliction has varying complexity depending on the configuration, using \cref{algo:mbd} directly is insufficient, as it requires a predefined number of denoising steps $N$ and lacks the anytime property. Therefore, we propose performing the denoising process iteratively, using \cref{algo:mbd} with a fixed small $N$ several times.

As shown in \cref{algo:incremental}, the crux here is that instead of simply using MBD to synthesize control trajectories from scratch, we synthesize a deformation control vector $\Delta U \in \mathcal{U}^{nH}$ for the previous iteration's trajectory.
In other words, we optimize $\Delta U$ via denoising to update the control trajectory in a solution $\tau$ as $\tau.u \leftarrow \tau.u + \Delta U$.
The samples in \cref{algo:mbd} now become deformation vectors rather than control trajectories, and the reward for a sampled deformation $\Delta U_\text{sample}$ is computed by rolling out $\tau.u + \Delta U_\text{sample}$, where $\tau$ is the trajectory obtained from the previous iteration and is fixed to serve as the base trajectory when denoising $\Delta U$ at current iteration.
The rationale is that each denoising process guides $\tau$ towards the target distribution $p_0$ and thus the amount of deformation becomes smaller and smaller over iterations.
This makes it an \emph{anytime} planning algorithm; over time, we can expect the solutions to get better and better.
The decision of when to stop depends on the applications and user requirements.
Some may stop refining when a feasible solution is reached, others may stop at the planning deadline.

The validity of this iterative denoising comes from MBD with the reverse process $p_{i-1|i}(\cdot)$ approximated by on-the-fly Monte Carlo score estimation.
Traditional diffusion models with neural networks that learn from data cannot cope with the shift in the target distribution of deformations during successive iterations.
Further, we observe empirically that initializing $\Delta U_N$ with a sample from a standard Gaussian distribution does not produce a steady refinement over the iterations, and instead, our implementation uses a zero vector as $\Delta U_N$.
Our hypothesis is that introducing a zero-mean inductive bias \textit{helps} the denoising process approximate the optimal deformation quicker, while adding large deformations to the control trajectory is more likely to cause dramatic changes to the position trajectory.
{
\begin{algorithm}[t!]
\caption{Iterative Denoising}
\label{algo:incremental}
\begin{algorithmic}[1]
\small
\State Initialize $\tau$
\While{not interrupted}
\State Get a deformation control vector $\Delta U$ using \cref{algo:mbd} and $\tau$
\State $\tau \leftarrow \texttt{rollout}\left( \mathcal{S}, \tau.u + \Delta U \right)$
\EndWhile
\State \Return $\tau$
\end{algorithmic}
\end{algorithm}
}

\section{Evaluations}
We evaluate the performance of our trajectory optimization method qualitatively and quantitatively through a variety of metrics.
A key feature of our method is the ability to use a variety of kinodynamic models, and thus we show evaluations on three types of systems:
\begin{itemize}[leftmargin=*]
    \item \textbf{Differential Drive} mixed-order integrator system that represents wheeled robots with
    state $x=[p_x, p_y, \theta, v]^\top$,
    control $u = [\omega, a]^\top$,
    and
    $f_{\mathrm{dyn}}(x, u) = [v\cos\theta, v\sin\theta,\omega, a]^T$;
    \item \textbf{2D Holonomic} double-integrator system that represents ground robots with
    state $x=[p_x, p_y, v_x, v_y]^\top$, control $u=[a_x, a_y]^\top$, and $f_{\mathrm{dyn}}(x, u) = [v_x, v_y, a_x, a_y]^T$;
    \item \textbf{3D Holonomic} double-integrator system, which is a 3D version of the 2D case.
\end{itemize}
All evaluations are carried out on a laptop PC with an Intel Core i9-13900HX CPU, equipped with an NVIDIA RTX 4080 GPU.
Our implementation is based on~\cite{pan2024model}, coded in Python with \texttt{JAX}\cite{jax2018github} library to streamline GPU acceleration.
We use planning horizon $H = 100$ with $\Delta t = 0.1$
and $M=2048$ trajectory rollouts to approximate gradients (scores).

\subsection{Qualitative Analysis}
\Cref{fig:denoising} depicts a qualitative investigation into the intermediate steps of the denoising process for each of these systems.
The problem considers a navigation task involving 8 and 16 robots placed on a circle (sphere, in 3D) with their respective destinations placed at antipodal locations.
Such scenarios have served as typical benchmarks in the multi-robot planning literature, as they always force planners to make non-trivial deconfliction efforts.
The snapshots in \cref{fig:denoising} depict the refinement across different denoising steps, i.e., over $i$ in \cref{algo:mbd}, and iterations of \cref{algo:incremental}.
The figure illustrates that the trajectories converge to their targets and are collision-free as the denoised steps increase, despite its enormous number of optimization variables, i.e., $d_u\times H\times n$.
As seen in the 2D holonomic case, which has $3200$ variables, iterative deformation optimization then compensates for incomplete trajectories and further improves solution quality by escaping local minima, by resetting the diffusion noise scheduler.

\subsection{Quantitative Evaluations: Sensitivity}
{
\newcommand{\entry}[3]{
  \node[] at (\xsize * #1, #2) {
    \includegraphics[width=0.33\linewidth,height=0.25\linewidth]{images/#3}
  };
}

\begin{figure}
\centering
\begin{tikzpicture}
  %
  \begin{scope}[xshift=0,yshift=0]
  \node[] at (1.50, 1.4) {\scriptsize \textbf{Differential Drive}};
  \node[] at (3.85, 1.4) {\scriptsize \textbf{2D Holonomic}};
  \node[] at (6.15, 1.4) {\scriptsize \textbf{3D Holonomic}};
  \node[] at (3.7, -4.1) {\small denoising steps ($N$)};
  \node[rotate=90] at (-0.2, -1.0) {\small iterations};
  \node[rotate=90] at (-0.5, -0.0) {\scriptsize 10 robots};
  \node[rotate=90] at (-0.5, -2.3) {\scriptsize 16 robots};
  \node[anchor=west] at (0, 0) {\includegraphics[width=0.9\linewidth]{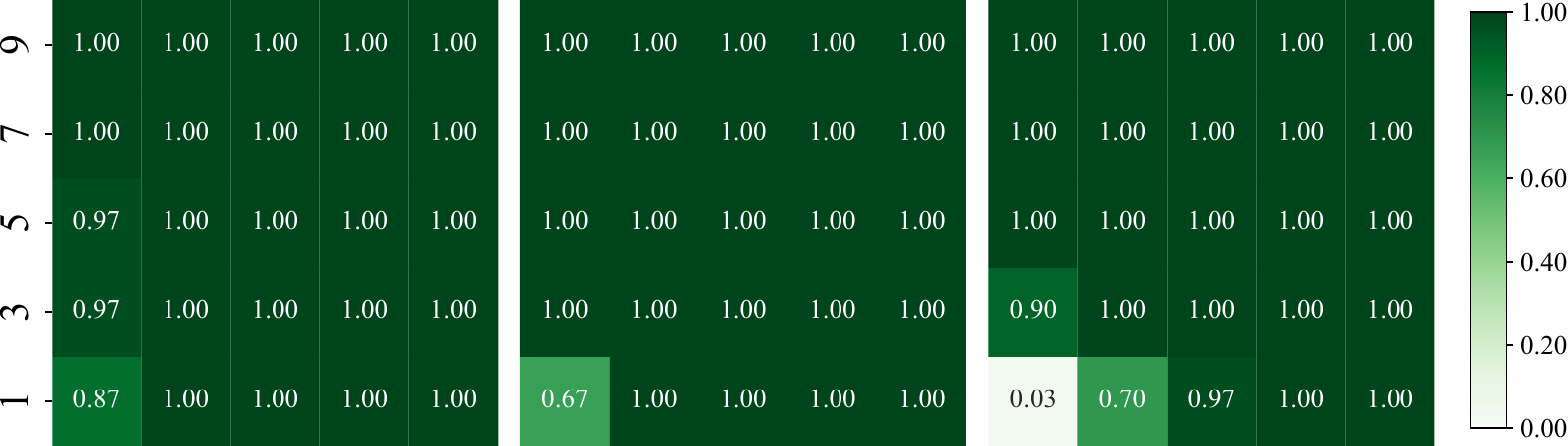}};
  \node[anchor=west] at (0, -2.6) {\includegraphics[width=0.9\linewidth]{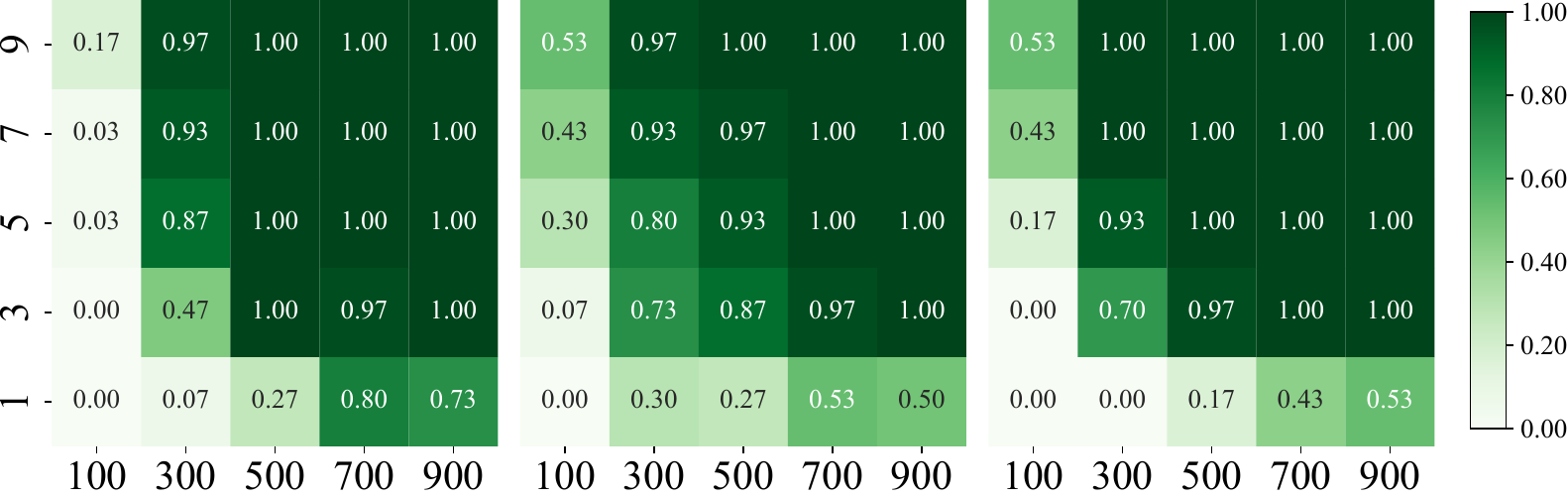}};
  \end{scope}
  %
  %
\end{tikzpicture}
\caption{
Sensitivity of the success rate to the number of diffusion steps and deformation iterations.
}
\label{fig:sensitivity}
\end{figure}
}

We now evaluate the performance of our iterative denoising process in terms of its success rate over the number of iterations and the number of denoising steps.
A solution is ``successful'' when conditions in \cref{eqn:problem-conditions} are met, i.e., feasible, conflict-free trajectories are found for all robots.
This evaluation is crucial from a practical perspective, since the total number of rollout operations, which directly affects the wall time, remains the same for $3\,\mathrm{iters}\times100\,\mathrm{steps}$ and
$1\,\mathrm{iters}\times300\,\mathrm{steps}$, even though these can produce some differences based on the difficulty of the problem.

\Cref{fig:sensitivity} presents an overview of success rates for 10 and 16 robot problems with the three dynamics models, averaged over 30 runs with different initial seeds.
We observe that several of the 10-robot cases are solved even with a single denoising iteration.
For 16 robots, we see some variability, and the results indicate that either increasing the denoising steps ($N$) or the number of iterations improves the success rate.
Nevertheless, in the following quantitative evaluations, we set $N=100$ to provide a more general setup while comparing against baselines. 
This choice keeps the computational time per iteration low, enabling more flexible termination of the algorithm.

\subsection{Quantitative Evaluations: Baselines} \label{sec:baselines}
{
\newcommand{\entry}[3]{
  \node[] at (\xsize * #1, #2) {
    \includegraphics[width=0.3\linewidth]{images/#3}
  };
}

\begin{figure}
\vspace{1pt}
\centering
\begin{tikzpicture}
  \scriptsize
  \tikzmath{
    \xsize = 2.75;
    \ya = 0;
    \yb = -5.0;
    \yc = -2.5;
  }
  \node[] at (\xsize * 0, 1.25) {\textbf{Differential Drive}};
  \node[] at (\xsize * 1, 1.25) {\textbf{2D Holonomic}};
  \node[] at (\xsize * 2, 1.25) {\textbf{3D Holonomic}};
  %
  \entry{0}{\ya}{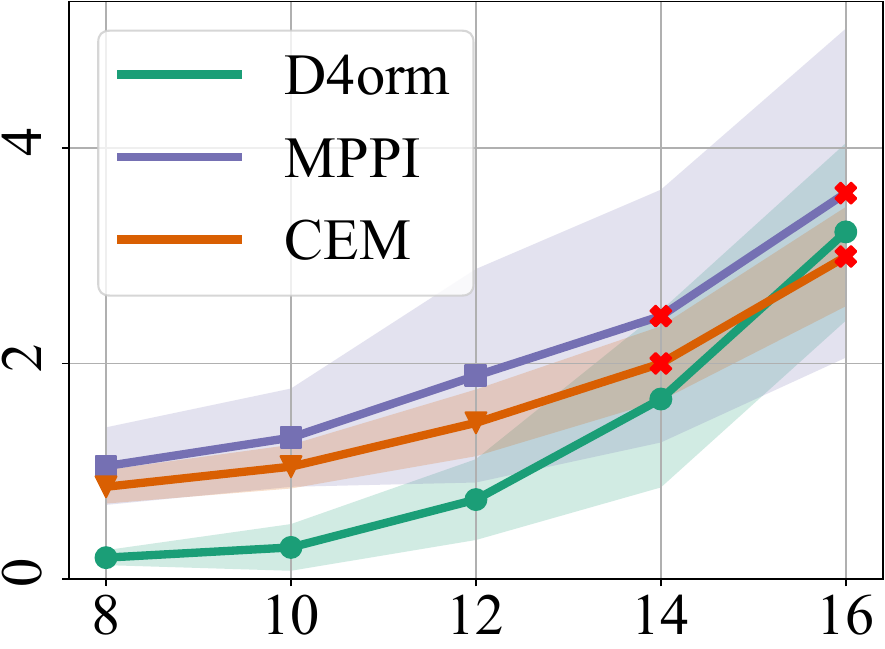}
  \entry{1}{\ya}{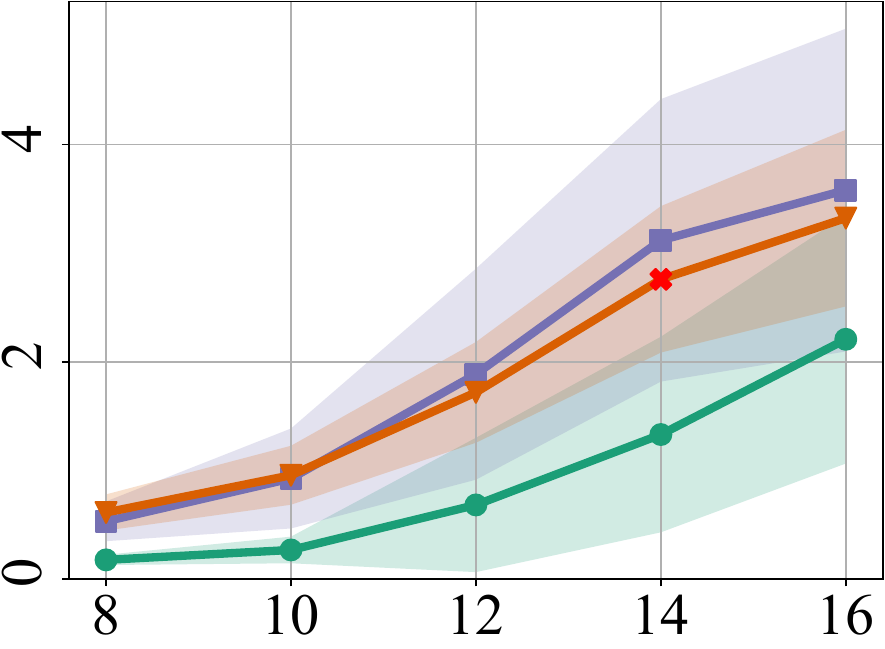}
  \entry{2}{\ya}{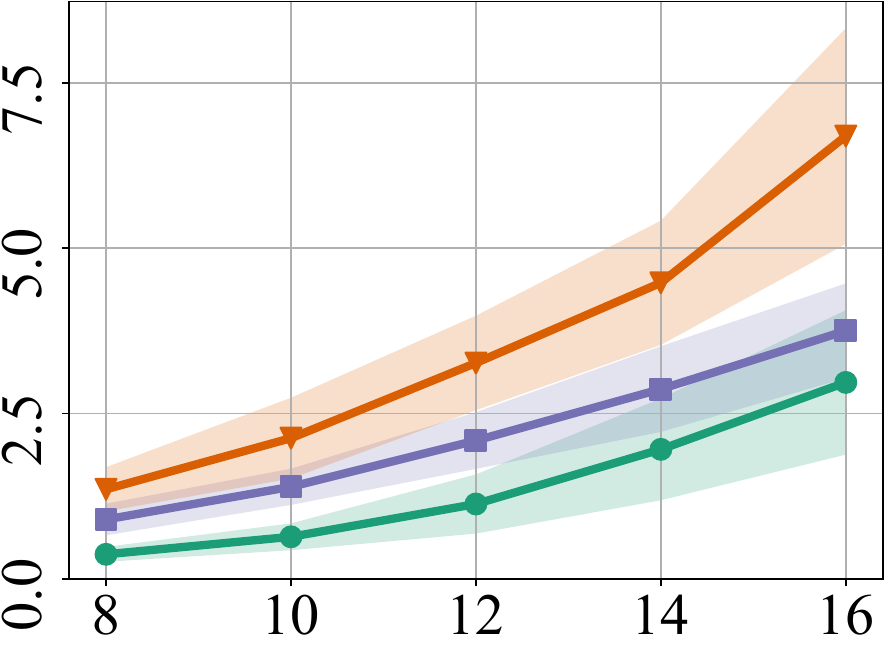}
  \node[rotate=90] at (-\xsize * 0.5 - 0.2, \ya) {$\leftarrow$ runtime [s]};
  \node[] at (\xsize * 1, \ya-1.2) {number of robots (\textcolor{red}{$\times$}: exclude failed runs, $\bullet$: all runs succeeded)};
  %
  \entry{0}{\yb}{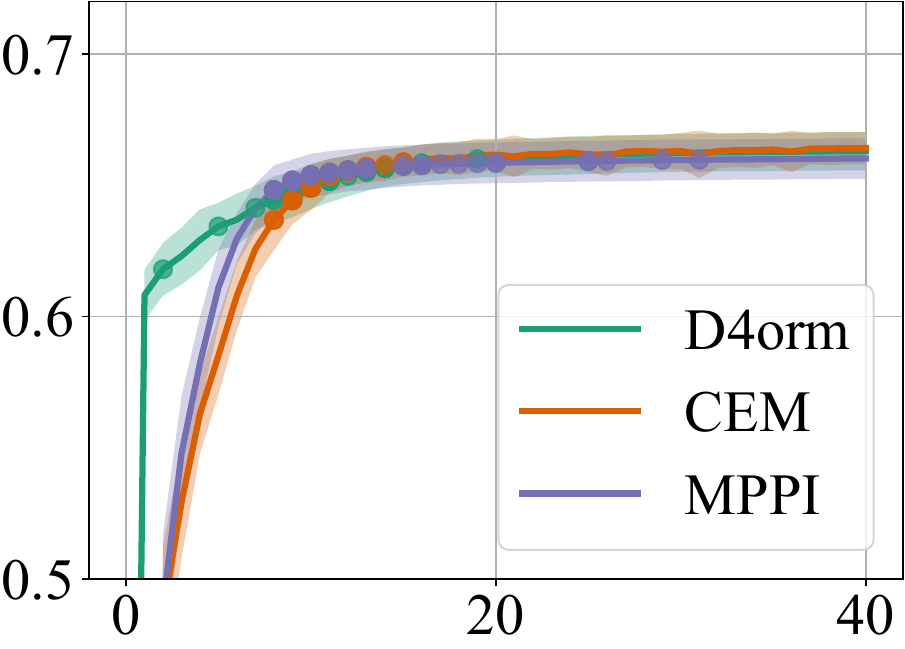}
  \entry{1}{\yb}{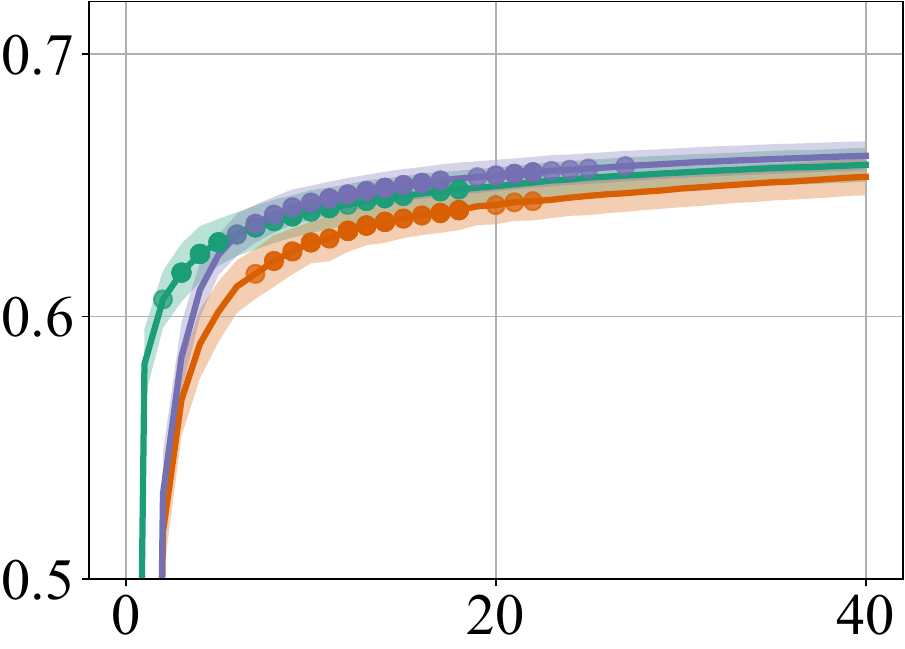}
  \entry{2}{\yb}{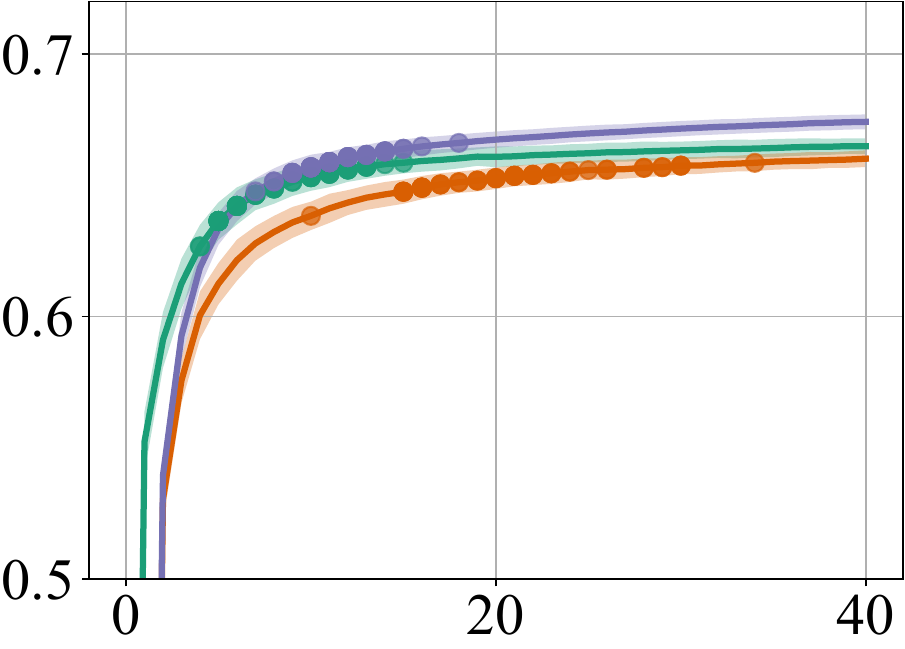}
  \node[rotate=90] at (-\xsize * 0.5 - 0.2, \yb) {reward $\rightarrow$};
  \node[] at (\xsize * 1, \yb-1.2) {number of total steps ($\times 10 ^ 2$)};
  %
  \entry{0}{\yc}{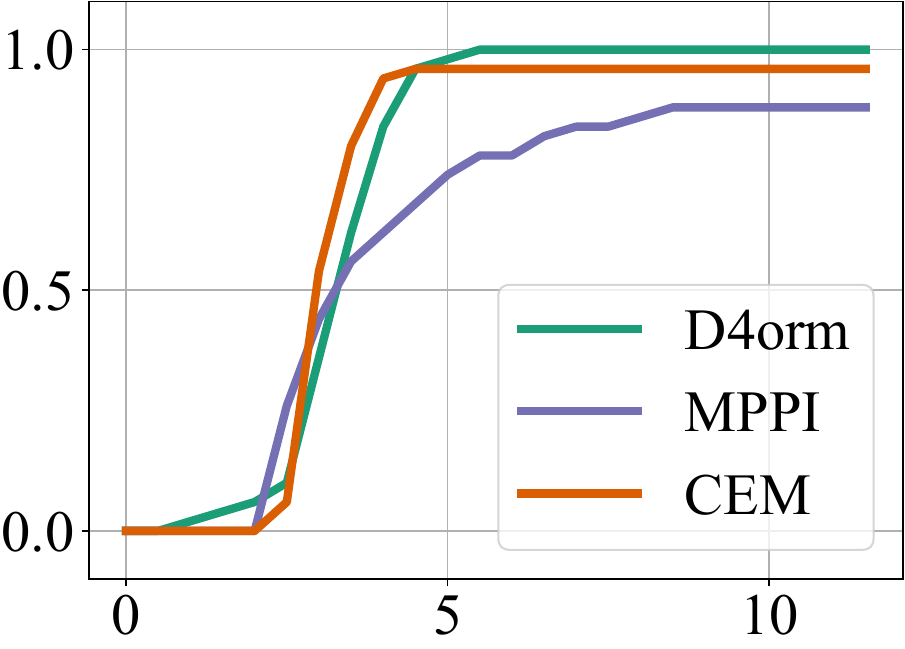}
  \entry{1}{\yc}{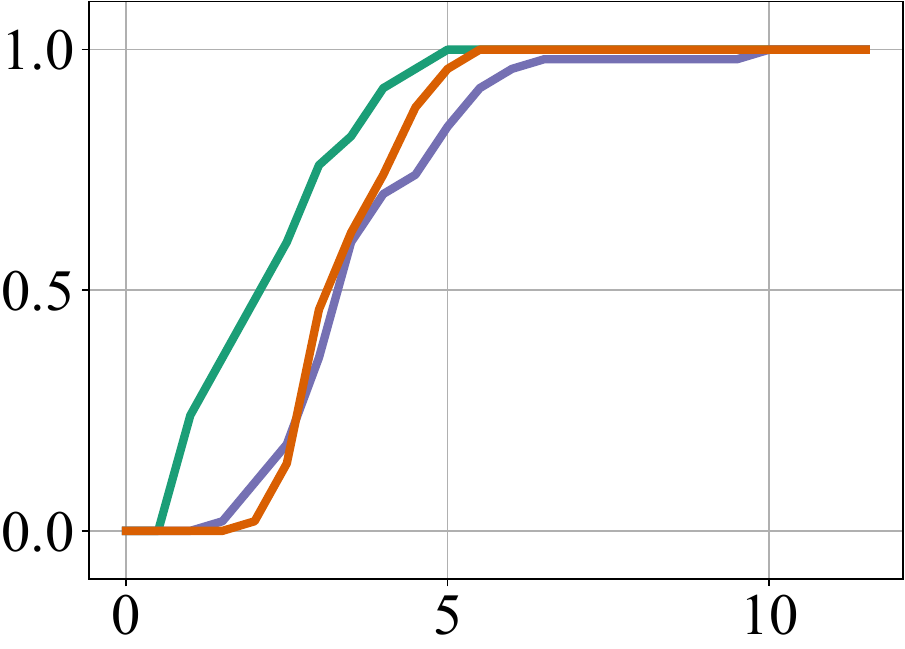}
  \entry{2}{\yc}{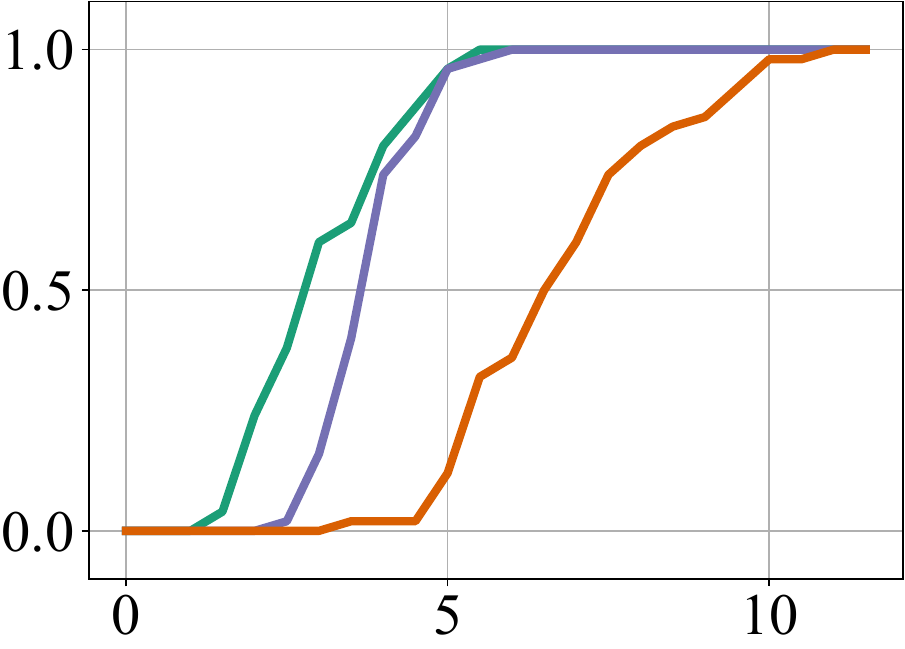}
  \node[rotate=90] at (-\xsize * 0.5 - 0.2, \yc) {success rate $\rightarrow$};
  \node[] at (\xsize * 1, \yc-1.2) {planning time [s]};
\end{tikzpicture}
\caption{
Empirical results against baseline methods.
The upper figures show the time required by each method to find initial feasible solutions over different numbers of agents.
The middle ones show the planner's ability to find a solution in the given time, focusing on the case of 16 robots.
The bottom ones show how quickly each method finds plausible solutions, along with the total number of steps ($\mathrm{iters} \times N$ for \textsc{D4orm}), using 16 robots each.
Instance-wise, when to find the initial solution is marked with $\bullet$.
}
\label{fig:results}
\end{figure}
}

Next, we contrast our method against two highly effective and competitive baseline methods that belong to the sampling-based optimization category: model-predictive path integral (MPPI)~\cite{williams2018information} and Cross Entropy Method (CEM)~\cite{botev2013cross}. For evaluation, these methods are adapted to perform sampling in the joint control space.
We evaluate three key metrics of interest:
\textit{runtime}, to quantify scalability against varying team sizes by measuring clock time for computing successful solutions,
\textit{reward}, to quantify solution quality according to \cref{eq:objective} when planning is allowed to continue for longer, and,
\textit{success rate}, as defined previously, but measured against a given planning deadline.

\Cref{fig:results} shows a complete overview of our comparisons over 50 runs.
We observe a noticeable improvement in \textbf{runtime} and success rate for our method against \textit{all} baselines for \textit{all} robot models and for \textit{almost all} team sizes.
In particular, the \textbf{success rate} plots (middle row) show that for the 2D holonomic case, our approach can always retrieve kinodynamically feasible and collision-free trajectories for 16-robots teams given a \SI{5}{s} deadline. The average runtime is lower, $\approx$ \SI{2}{s}, which is a near $2\times$ improvement over MPPI and CEM.
The \textbf{reward} plots at the bottom show the `anytime planning' nature using the number of steps, which serves as a metric for assessing runtime independently of computing environments.
Recall that the reward is an indicator for solution quality, and as such, all methods converge to similar reward values given sufficient time for refinement.
However, our proposed method shows two key advancements.
First, its reward curves show a steeper gradient during early steps, indicating that we obtain high-quality solutions faster.
Second, as indicated by the solid dots on the curves, our high-quality solutions are successful (i.e., conflict-free) earlier than other methods of similar quality.
This is most noticeable in the 3D holonomic case, where our method offers a dramatic $\approx\SI{60}{\percent}$ reduction in the total steps required compared to CEM.
These improvements are owing to the ability of diffusion denoising to capture complicated, multimodal reward distributions in high-dimensional spaces.

\subsection{Real-World Deployment}
As evidence of safety and feasibility, we deploy trajectories generated by denoising in a zero-shot manner for a team of eight multirotors~\cite{woo2025sanity}.
\cref{fig:denoising} presents the deployment snapshot as well as the minimum inter-robot distance measured over several repeated executions beyond one minute, demonstrating that the robots maintain a safe distance.
The video is available in the supplementary materials, which also include demonstrations of more advanced planning scenarios with various obstacle densities, a larger number of robots, and random target assignments.

\section{Discussion}
We studied a multi-robot trajectory optimization framework with diffusion denoising.
Given robot dynamics and a reward function, the framework iteratively applies a deformation vector to the current candidate solution, which is guided by a Monte Carlo score approximation.
Despite the high dimensionality of deconfliction problems, our evaluations demonstrate that this simple process successfully retrieves collision-free trajectories for the entire team within reasonable deadlines.
There are technical considerations surrounding the framework's reliance on numerous and long-horizon rollouts, which is currently the most expensive subprocess.
This necessitates using GPUs to speed up gradient approximation.
The framework can be applied to non-spherical shapes at the cost of more expensive inter-robot collision checking.
Generally, improving sampling efficiency is thus important.
Nevertheless, our future work is addressing several directions unexplored in this paper. These include handling obstacle-rich settings, and heterogeneous teams.

\bibliographystyle{sty/IEEEtran}
\bibliography{sty/ref-macro,ref}
{
\newcommand{\entry}[4]{
\node[anchor=west] at (#1 * \xstep, 0) {\includegraphics[width=0.19\linewidth]{images_supp/#3}};
\node[anchor=west,inner sep=0,draw] at (#1 * \xstep + 0.51, 0.56) {\includegraphics[width=0.07\linewidth,height=0.07\linewidth,trim={30 30 5 5},clip]{images_supp/#2}};
\node[anchor=center] at (#1 * \xstep + 2.0, 1.4) {\small{#4}};
}

{
\newcommand{\entrysimple}[4]{
\node[anchor=west] at (#1 * \xstep, 0) {\includegraphics[width=0.19\linewidth]{images_supp/#3}};
}
\begin{figure*}[tp]
  \tikzmath{
    \xstep = 3.45;
  }
\centering
\begin{tikzpicture}
  \entry{0}{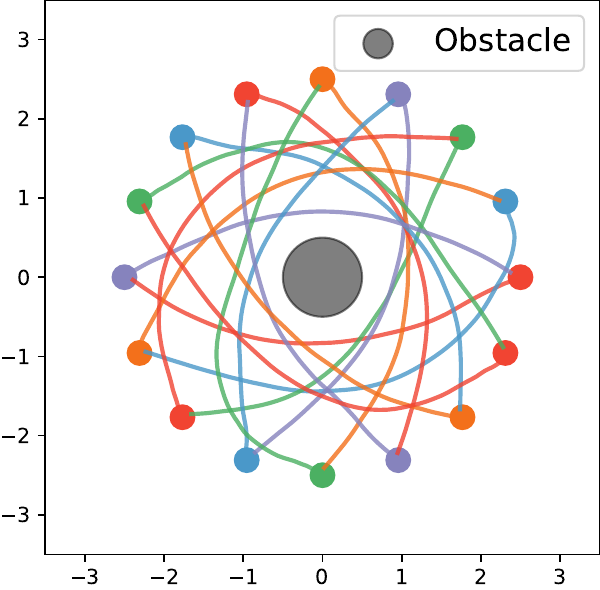}{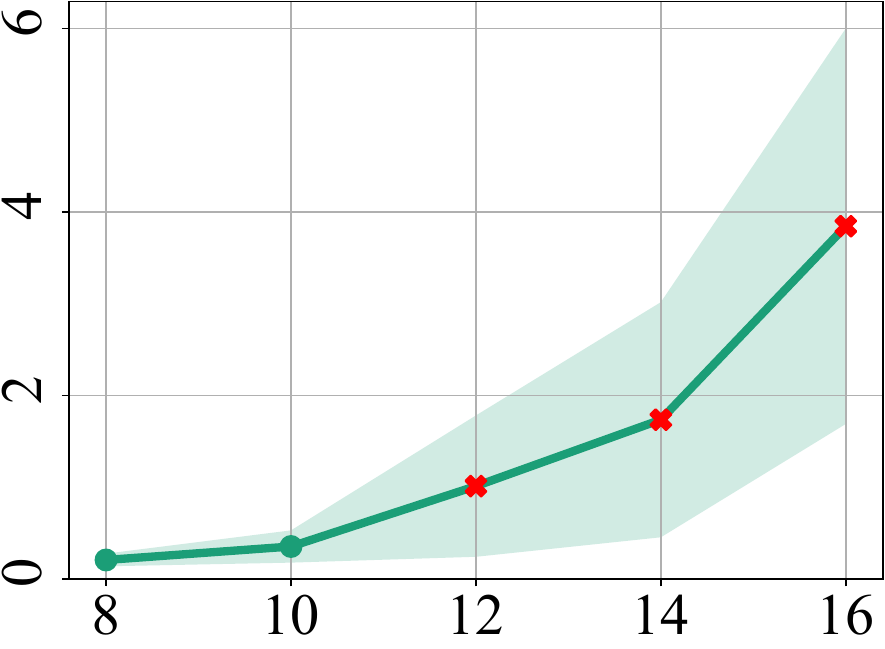}{1 large obs.}
  \entry{1}{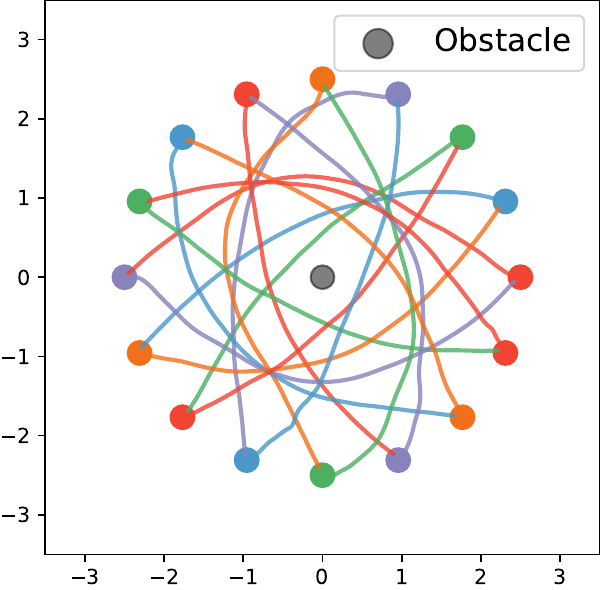}{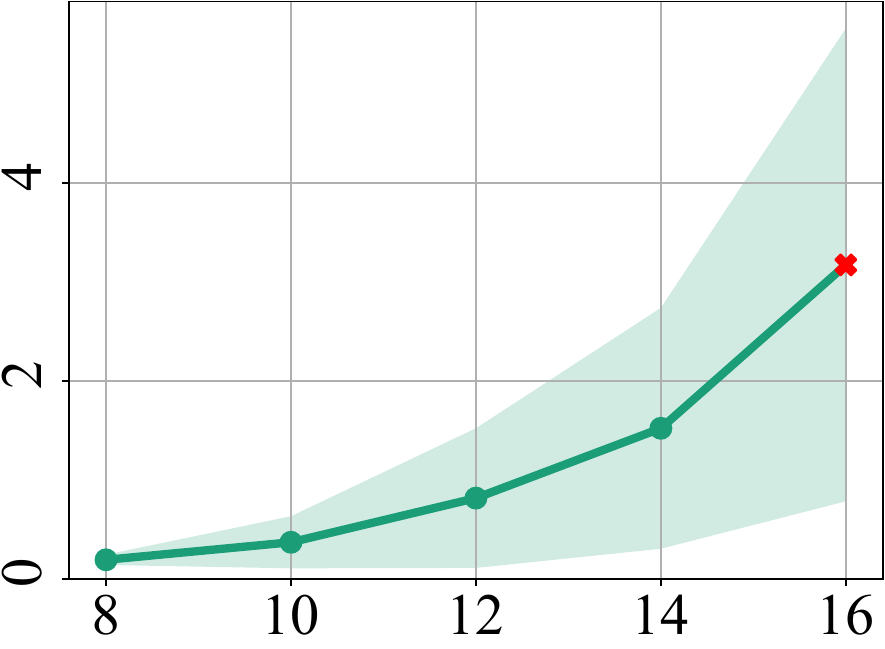}{1 small obs.}
  \entry{2}{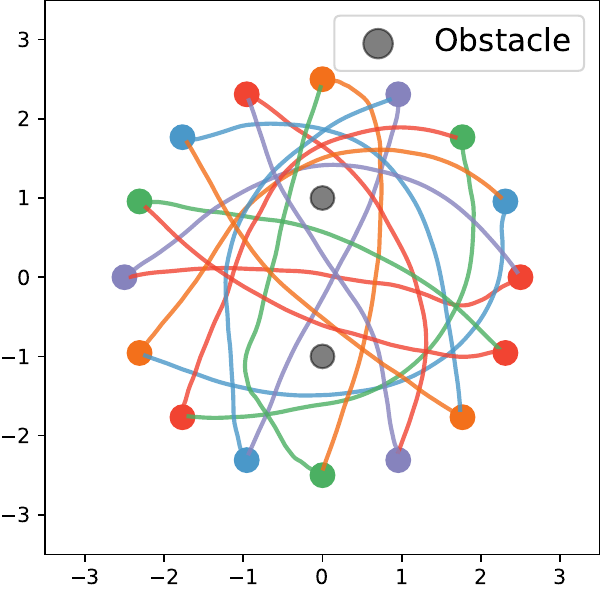}{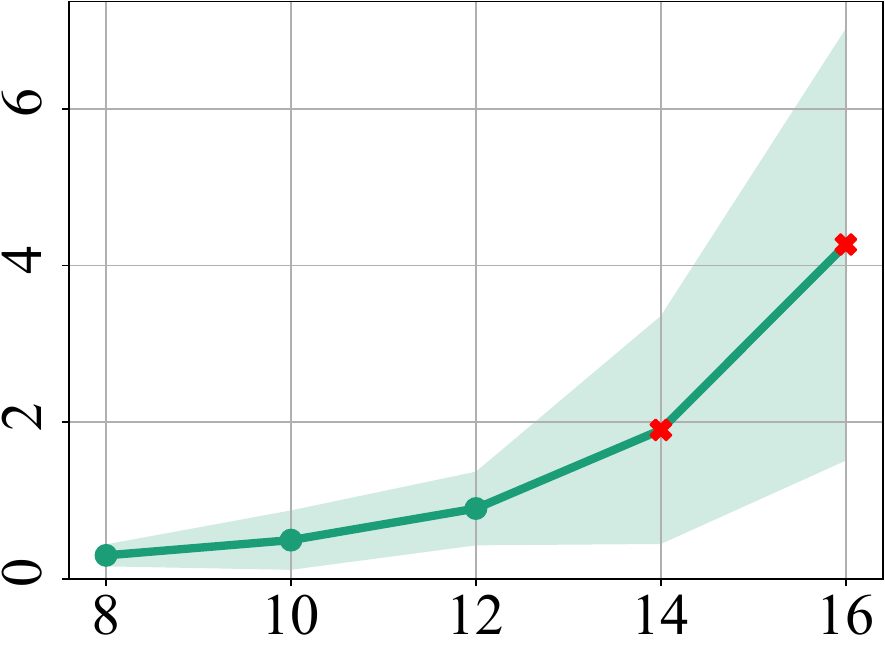}{2 obs.}
  \entry{3}{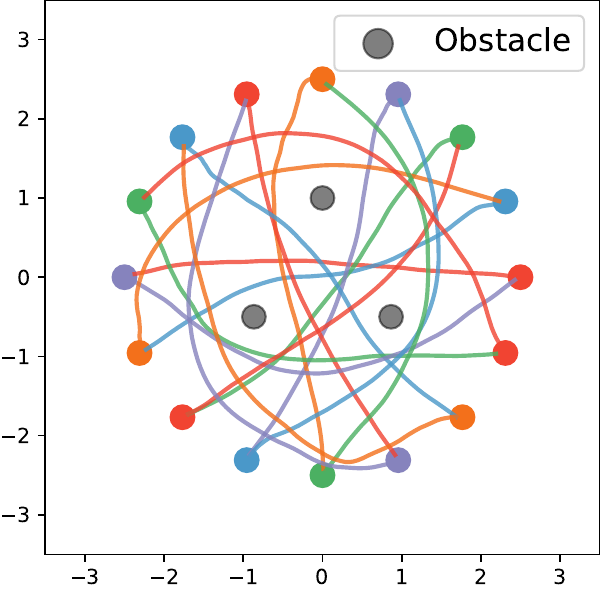}{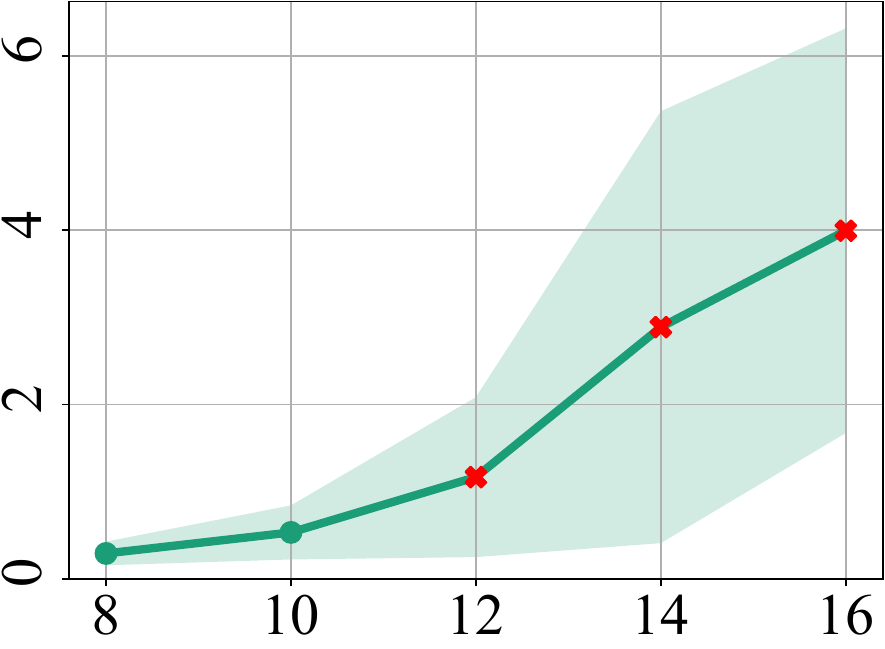}{3 obs.}
  \entry{4}{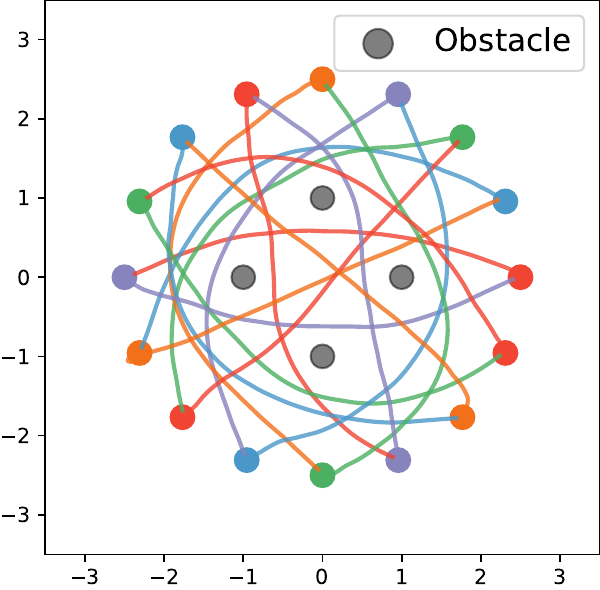}{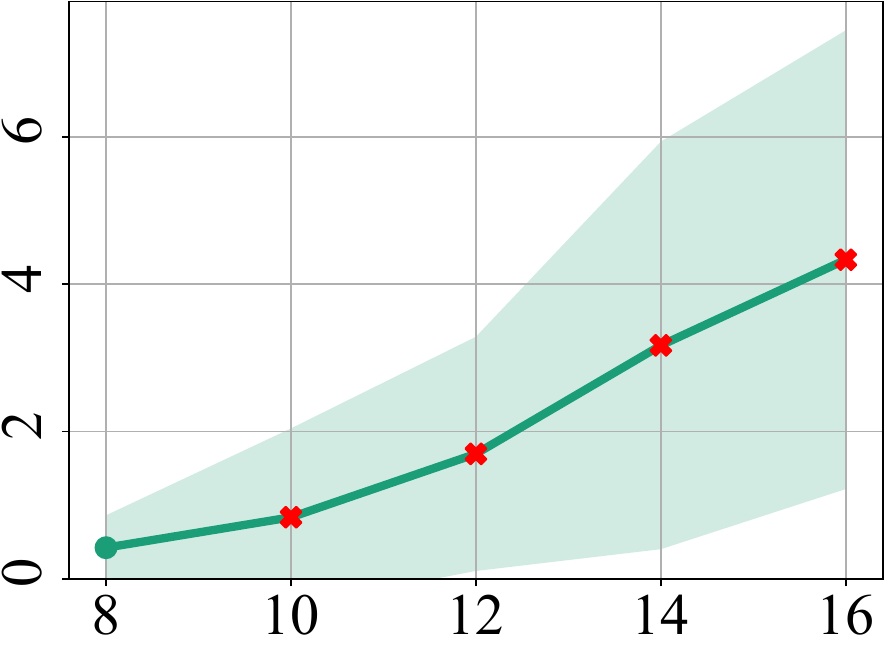}{4 obs.}

  \node[anchor=center,rotate=90] at (-0.2, 0.2) {\small runtime [s]};
\end{tikzpicture}

\begin{tikzpicture}
  \entrysimple{0}{16a_1obs_big}{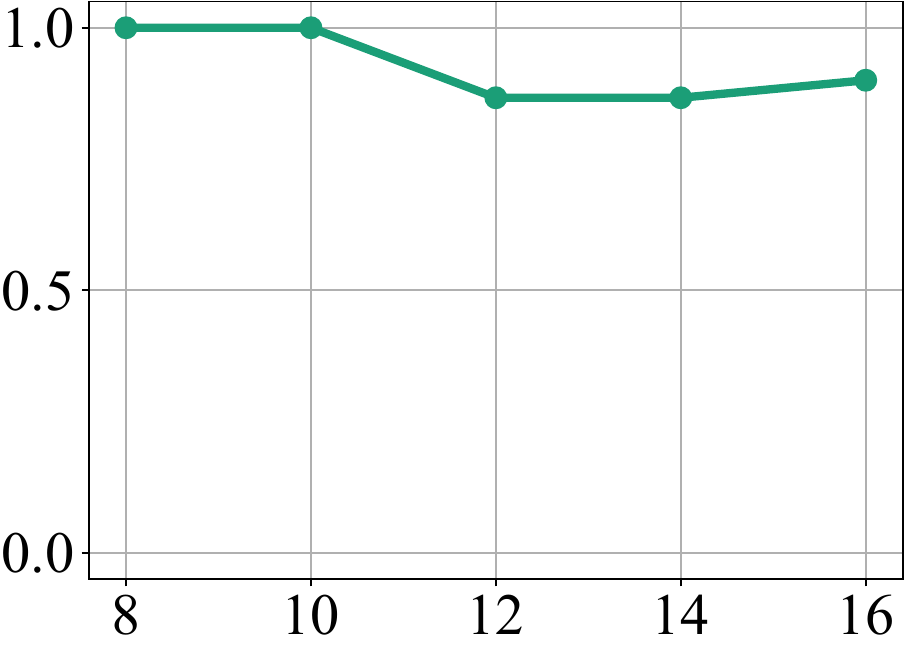}{1 large obs.}
  \entrysimple{1}{16a_1obs}{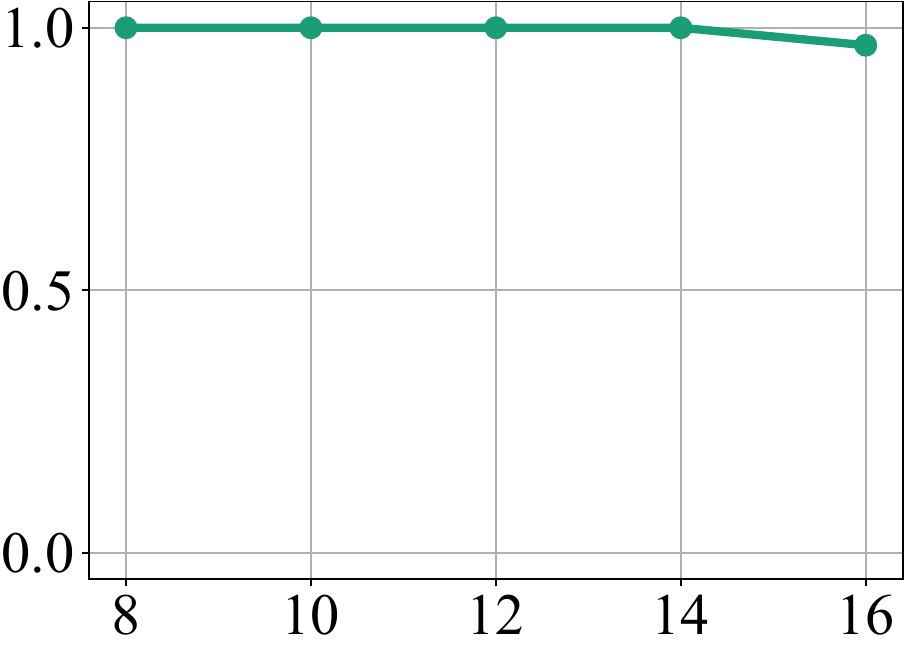}{1 small obs.}
  \entrysimple{2}{16a_2obs}{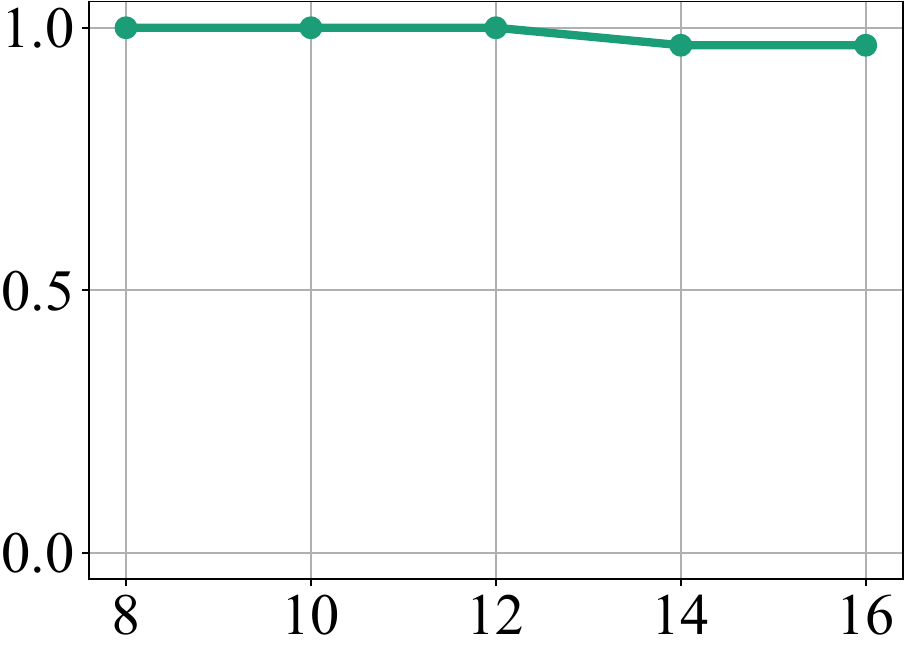}{2 obs.}
  \entrysimple{3}{16a_3obs}{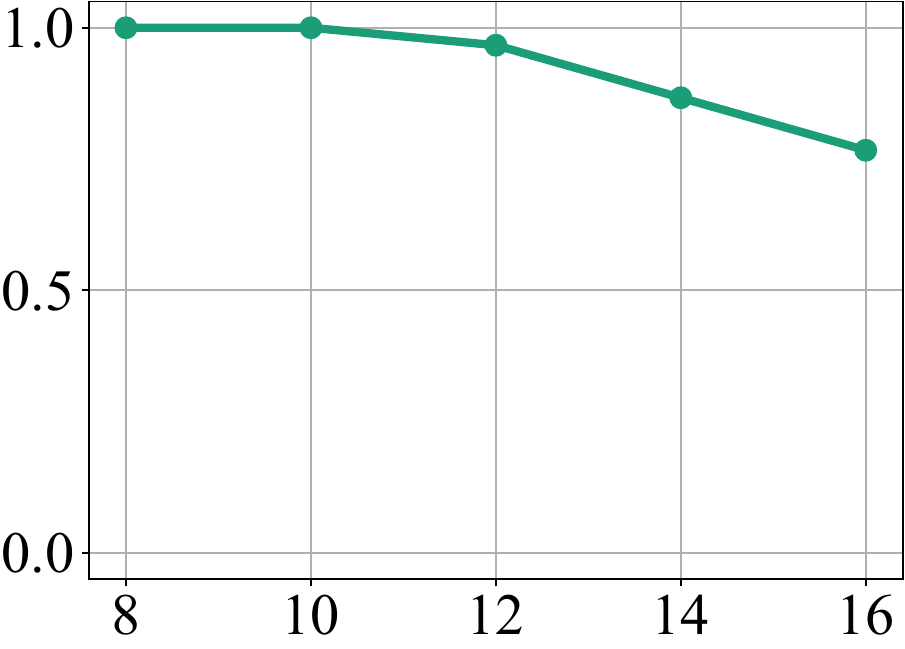}{3 obs.}
  \entrysimple{4}{16a_4obs}{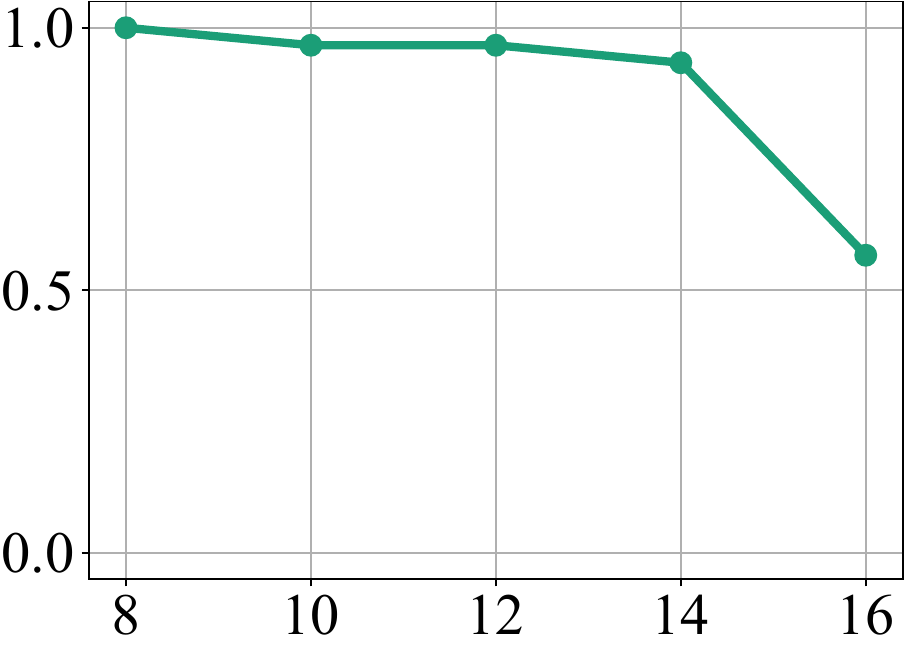}{4 obs.}

  \node[anchor=center,rotate=90] at (-0.2, 0.2) {\small success rate};
  \node[anchor=center] at (9.0, -1.4) {\small number of robots};
\end{tikzpicture}

\caption{
Antipodal navigation scenarios for 16 robots with varying numbers of obstacles. Top row: The computation time for finding initial solutions is provided, averaged across 30 runs, excluding any failed runs. Bottom row: The success rate over 30 runs, with a maximum of 50 iterations.
}
\label{fig:obstacles}
\end{figure*}
}

\appendix
We now present some additional studies with \textsc{D4orm} to showcase our ability to handle general target assignment scenarios (not only antipodal points on a circle/sphere), scaling up to more number of robots, and handling more complex workspaces that contain obstacles.
\newline{}

\medskip\noindent\textbf{Generic Target Assignment.}
To validate that the success of our proposed method does not depend on exploiting biases or patterns in the problem, we test it on $30$ random configurations with $\{8\ldots 16\}$ 2D holonomic robots.
The environments have random initial and target positions, all generated within a square of size \( D \times D \), with \( D \) as the diameter of the circle with antipodal points used in previous experiments.
\cref{fig:time_cmp_random_16} (left) shows a solution instance for 16 robots.
We also observe in \cref{fig:time_cmp_random_16} (right) that the method works reliably for all environments and is generally faster; the average time required to generate the first feasible solution for 16 robots is about \SI{1}{\second}, which is less than the circle with antipodal points scenario ($\approx\SI{2}{s}$).
This is because some of the trajectories do not interfere with each other due to geometric relationships in the random scenario, and are therefore often much easier to deconflict.

\begin{figure}[htbp]
    \begin{tikzpicture}
        \node[] at (0,0)
        {   
            \includegraphics[width=0.37\columnwidth]
                {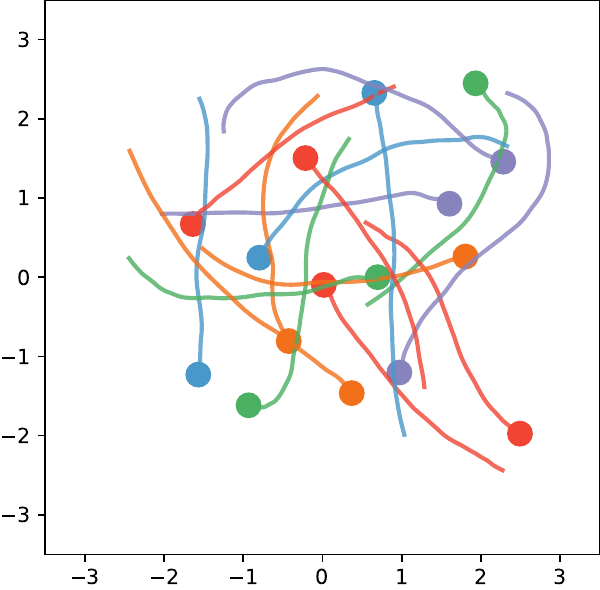}
            \hspace{0.5cm}
            \includegraphics[width=0.49\columnwidth]
                {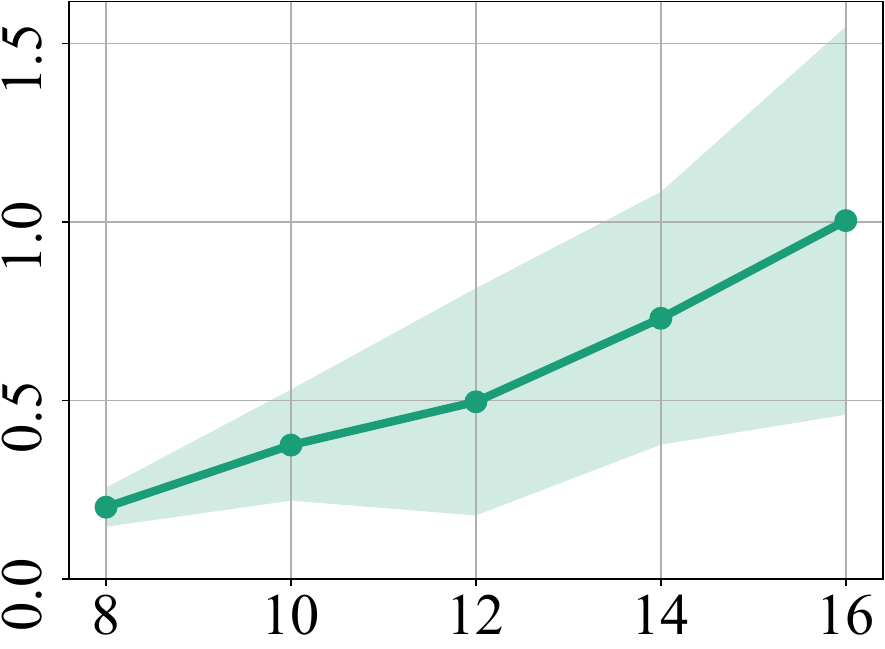}
        };
        \node[] at (2.2, -1.8)
        {
            {\small \#robots}
        };
        \node[rotate=90] at (-0.4,0)
        {
            {\small runtime [s]}
        };
        \draw[fill=gray!50,draw=none] (-3.8,-1.3)
            rectangle ++(1.8,0.35)
            node[pos=.5]
            {{\small\texttt{\#robots=16}}};
    \end{tikzpicture}
    \caption{
    A solution instance for a scenario with random targets, and time required by \textsc{D4orm} to find initial feasible solutions.
    }
    \label{fig:time_cmp_random_16}
\end{figure}

\medskip\noindent\textbf{Presence of Obstacles.}
We also tested our method in environments with obstacles, which can be easily addressed by a slight modification of the reward function of \cref{eqn:generic-reward}.
In particular, for each state in the trajectories, we add a binary penalty term for whether the robot collides with any obstacle.
\Cref{fig:obstacles} shows that \textsc{D4orm} is capable of handling obstacles.
Meanwhile, as the workspace becomes more complex due to the addition of obstacles, the time required to obtain the first feasible solution increases, and \textsc{D4orm} also encounters failures where a few robots are sacrificed for collision in order to achieve a high team reward.
Dealing with obstacle-rich environments in a more systematic way (as opposed to a na\"{i}ve binary penalty) is an interesting future direction.


\end{document}